\documentclass[journal,twocolumn]{IEEEtran}
\IEEEoverridecommandlockouts
 
\AtBeginDocument{%
  \providecommand\BibTeX{{%
    \normalfont B\kern-0.5em{\scshape i\kern-0.25em b}\kern-0.8em\TeX}}}

\usepackage{color,soul}
\usepackage{algorithmicx}
\usepackage{algorithm}
\usepackage{algpseudocode}
\usepackage{adjustbox}
\usepackage{comment}
\usepackage[utf8]{inputenc}
\usepackage{url} % For formatting URLs
\usepackage{hyperref} % For creating clickable links
\usepackage{enumitem} % For customizing enumerate labels
\usepackage{graphicx}
\usepackage{caption}
\usepackage{subcaption}
\usepackage{comment}
\usepackage{wrapfig}
\usepackage{booktabs}
\usepackage{balance}
\begin{document}

\title{
Thermal Vision: Pioneering Non-Invasive Temperature Tracking in Congested Spaces
}

\author{
\IEEEauthorblockN{Arijit Samal, Haroon R Lone \\
Department of EECS, Indian Institute of Science Education and Research Bhopal, India}% <-this % stops a space \\
%\IEEEauthorblockN{Anonymous Authors}% <-this % stops a space \\
\thanks{Corresponding author: Haroon R Lone (e-mail: haroon@iiserb.ac.in).\protect\\
 }% <-this % stops a space

\thanks{
Arijit Samal and Haroon R Lone are with the Department of Electrical Engineering and Computer Science,
IISER Bhopal, Madhya Pradesh, 462066, India (e-mail: arijits19@iiserb.ac.in; haroon@iiserb.ac.in)
}
\thanks{Digital Object Identifier 12.4209/LSENS.2011.0000000}
}
\maketitle
%\IEEEpeerreviewmaketitle

%\IEEEtitleabstractindextext{%
\begin{abstract}
 
 Non-invasive temperature monitoring of individuals plays a crucial role in identifying and isolating symptomatic individuals. Temperature monitoring becomes particularly vital in settings characterized by close human proximity, often referred to as \textit{dense settings}. However, existing research on non-invasive temperature estimation using thermal cameras has predominantly focused on \textit{sparse settings}. Unfortunately, the risk of disease transmission is significantly higher in dense settings like movie theaters or classrooms. Consequently, there is an urgent need to develop robust temperature estimation methods tailored explicitly for dense settings.

Our study proposes a non-invasive temperature estimation system that combines a thermal camera with an edge device. Our system employs YOLO models for face detection and utilizes a regression framework for temperature estimation. 
We evaluated the system on a diverse dataset collected in dense and sparse settings. 
Our proposed face detection model achieves an impressive mAP score of over 94 in both in-dataset and cross-dataset evaluations. Furthermore, the regression framework demonstrates remarkable performance with a mean square error of 0.18$^{\circ}$C and an impressive $R^2$ score of 0.96.
Our experiments' results highlight the developed system's effectiveness, positioning it as a promising solution for continuous temperature monitoring in real-world applications. With this paper, we release our dataset and programming code publicly. 
\end{abstract}

\maketitle

\section{Introduction}

Non-invasive temperature monitoring is essential for identifying and isolating symptomatic individuals to prevent the spread of disease \cite{STANIC2025115879,10014654}. During the COVID-19 pandemic, thermal cameras were widely adopted and often installed at building entrances to provide a one-time temperature reading~\cite{brzezinski2021automated,quilty2020effectiveness,zhou2020clinical}. These cameras were commonly deployed in high-traffic locations such as airports, universities, and retail outlets\footnote{\url{https://www.flir.in/instruments/public-safety/environmental-health-and-safety}}. While effective in detecting symptomatic individuals at entry points, this approach has limitations. Disease-related symptoms can develop at any time during the workday, making it necessary to monitor building occupants continuously. Solely relying on entry-level temperature screenings is insufficient for identifying symptomatic individuals and mitigating the risk of disease transmission within buildings. To address this gap, additional measures, such as wearable temperature monitoring devices or periodic temperature checks, should be considered. Implementing such measures would enable seamless, real-time monitoring and timely interventions. A comprehensive approach to temperature monitoring can significantly enhance disease prevention efforts, safeguarding the health and well-being of all building occupants. Recent advancements in robotic-assisted surgery \cite{alsajri2024future} have demonstrated how integrating real-time, precision-driven technologies into clinical workflows can substantially enhance patient outcomes. Inspired by these developments, our work extends similar principles to the domain of non-invasive temperature monitoring in congested spaces.

Temperature estimation can be performed in sparse or dense settings~\cite{li2019robust,ranjan2016thermalsense,wei2020low,ghassemi2018best,lin2019thermal,somboonkaew2020temperature}. Sparse settings refer to scenarios where a camera's field of view (FOV) captures only a few individuals, while dense settings involve multiple people in close proximity, engaging in activities such as sitting, standing, or walking. Most existing research on thermal image-based temperature estimation focuses on sparse settings. However, real-world environments—such as offices, classrooms, airports, movie theaters, and laboratories—often present dense settings where the risk of disease transmission is significantly higher. This underscores the need for continuous, non-invasive monitoring methods tailored to dense settings. Developing effective temperature estimation techniques for these complex scenarios is therefore critical.

This paper introduces a novel non-invasive temperature monitoring system for dense and sparse settings. The system integrates a FLIR Lepton 3.5 thermal camera with an embedded edge device (Raspberry Pi) for real-time temperature estimation. Our approach addresses the limitations of previous research through a robust, multifaceted pipeline. Live thermal data is processed in real-time on the edge device, where face detection is performed using the lightweight YOLOv11n model~\cite{glenn_jocher_2022_7347926}, and temperature estimation is conducted using a regression framework. The compact and efficient design ensures the system's adaptability to various embedded applications.

To enhance usability, we establish a versatile temperature monitoring pipeline capable of seamless adaptation to diverse scenarios, ensuring broad applicability. The system's training is supported by a curated thermal image dataset collected under various conditions, enhancing its robustness. During real-time monitoring, the system displays bounding boxes and estimated temperatures around detected faces in the live video stream, making it ideal for dynamic and real-time applications.The contributions of our work are as follows:
\begin{itemize}
    \item We have created a non-invasive embedded temperature monitoring system capable of continuously displaying people's temperatures within the camera's FOV. The system requires mounting on a wall and an external display to view real-time temperature readings. With this paper, we release the programming code publicly~\cite{indian-dataset}.
    \item We have thoroughly evaluated the system in dense and sparse settings, ensuring its effectiveness. We have conducted evaluations within the dataset and across different datasets to assess the system's robustness.
    \item We have collected a large-scale thermal image dataset with varying settings. In this paper, we publicly release the dataset, making it available to the research community~\cite{indian-dataset}.
\end{itemize}

\section{Related Work}
Existing research on temperature estimation encounters several significant challenges:
(i) \textit{High Cost and Scalability Issues:} Many studies utilize expensive thermal cameras and complex machine learning models, which limit scalability due to their high cost \cite{lin2019thermal, maguire2021thermal, cheong2014novel, chin2021infrared}.
(ii) \textit{Dual-Camera Dependencies:} Some approaches rely on separate regular RGB cameras for face detection and thermal cameras for temperature estimation. This dual-camera setup increases overall costs and may introduce inaccuracies, as the models are not specifically trained on thermal images \cite{hou2022low, mushahar2021human}.
(iii) \textit{Evaluation Approach:} Numerous studies use offline datasets for both training and testing, failing to evaluate their methods in real-world settings, which can lead to unrealistic performance expectations \cite{lin2019thermal, cheong2014novel, chin2021infrared, mushahar2021human, 9781417}.
(iv) \textit{Threshold-Based Approaches:} Fixed R, G, and B threshold-based methods are sometimes used, but these require careful calibration of thermal cameras to a black body. They are also sensitive to environmental factors, which can cause fluctuations in pixel values and reduce reliability \cite{rgb, mushahar2021human}.
(v) \textit{Complex 3D Head Models:} Certain methodologies employ intricate 3D head models to map facial features and estimate temperature. However, these processes are time-intensive and often result in mapping errors, leading to inaccurate temperature measurements \cite{hou2022low}.
(vi)  \textit{Limited Generalizability:} Many systems are designed for specific use cases, making them non-generalizable. Such systems perform poorly on unseen data and lack adaptability for other temperature monitoring applications \cite{guo2022development, sung2023image}. A high-level summary of the most relevant works can be found in Table \ref{tab:rel-works}.

\begin{table*}[!t]
  \caption{Summary of related works and comparison with our Work.}

  \centering
  \scriptsize
\hfill \break
    \begin{tabular}{p{1.5cm} p{5cm} p{4cm} p{4cm} p{1.2cm}}
    \toprule
        \textbf{Ref.} & \textbf{Objective} & \textbf{Camera} &  \textbf{Algorithm} & \textbf{Subjects}  \\
        \midrule
  \cite{lin2019thermal}  &A non-contact, continuous body temperature measurement system &  FLIR Lepton 2.5 thermal camera & Single-Shot-Multibox Detector and MobileNet architecture & Healthy\\
 \addlinespace[0.1cm]
  \cite{maguire2021thermal}  &  A thermal camera-based mass COVID screening system & High temperature detection system from Thales UK Limited & Calibrated with a blackbody source at typical skin temperatures & Clinical   \\
\addlinespace[0.1cm] 
  \cite{chin2021infrared}  & A rapid temperature screening system over long distances (2 m to 10 m) & FLIR E8-XT thermal camera & A thermal compensation model & Healthy \\
\addlinespace[0.1cm]
  \cite{hou2022low}&  A low-cost in-situ system for continuous
multi-person fever screening &  FLIR one pro & YOLO and custom temperature estimation model & Clinical \\
\addlinespace[0.1cm]

  \cite{mushahar2021human} &  A thermal imaging and screening system & AMG8833 IR $8\times8$ infrared thermal imager & YOLO and custom temperature estimation model & Healthy\\
  
    \textbf{Our work} &  A non-invasive, real-time temperature monitoring system designed for \textbf{densely populated environments.} &  FLIR Lepton 3.5 thermal camera   & Fine-tuned state-of-the-art (SOTA) YOLOv11 for thermal face detection in dense settings, combined with a custom regression framework for precise temperature estimation. Evaluated on both in-dataset and cross-dataset settings to ensure generalizability. & Healthy\\
    
  \bottomrule
    \end{tabular}
    \label{tab:rel-works}
\end{table*}

\section{Experimental Setup \& Data Collection}

The proposed system utilizes the FLIR Lepton 3.5 thermal camera module in combination with the Raspberry Pi 4B board for real-time thermal image acquisition and processing.

The FLIR Lepton 3.5 is a radiometric long-wave infrared (LWIR) camera capable of capturing precise temperature data for every pixel in its 160 $\times$ 120 resolution images. With a 57° field of view and a built-in shutter, it is compact and cost-effective, making it ideal for mobile and resource-constrained applications.

\begin{figure}

    \begin{center}
             \includegraphics[scale=0.18]{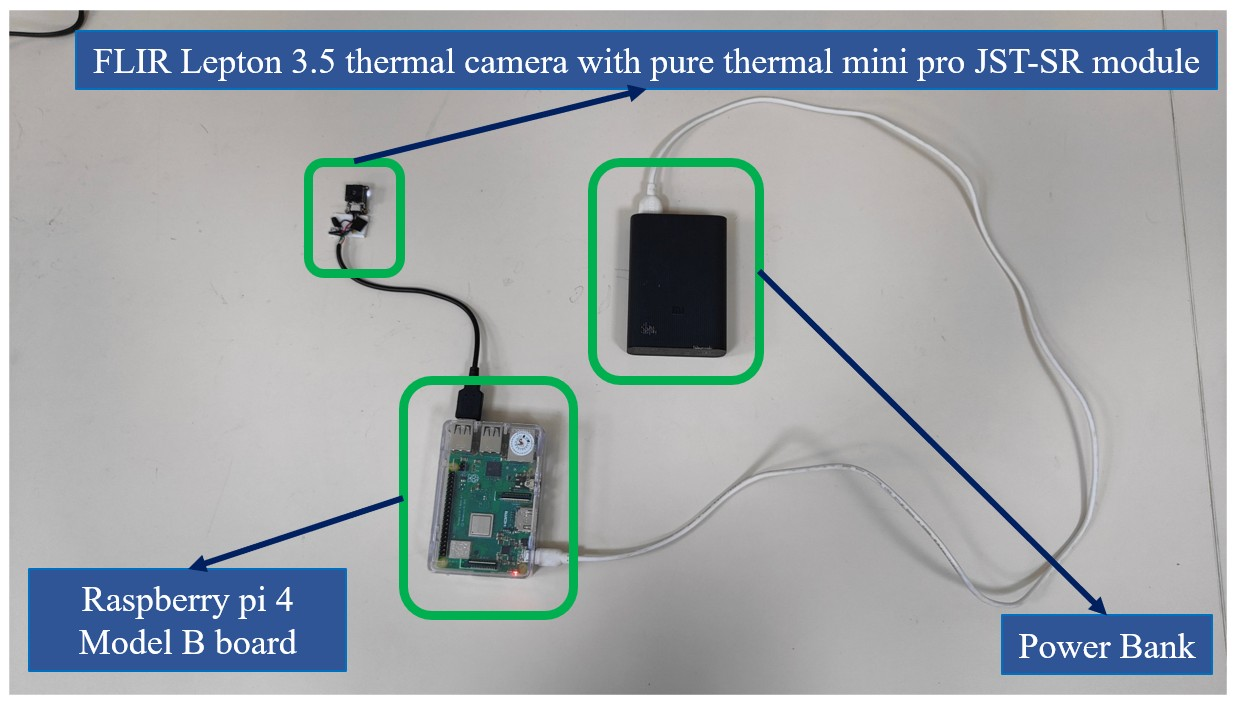}  
    \end{center}
        \caption{Experimental Setup}
        \label{fig:setup-figure}
%\end{wrapfigure}
\end{figure}

The Raspberry Pi 4B is the computational backbone, featuring a Broadcom BCM2711 quad-core Cortex-A72 processor running at 1.5 GHz, up to 8 GB of LPDDR4 RAM, and 32 GB of expandable microSD storage. Its connectivity options, including USB 3.0 and dual-band Wi-Fi, enable seamless data processing and integration with external devices.
Thermal images were collected using the FLIR Lepton 3.5 thermal camera connected to the Raspberry Pi 4B, powered by a portable power bank, as shown in Figure~\ref{fig:setup-figure}. Data acquisition was performed using the GuvcView library and custom Python scripts. A VNC Viewer was employed to establish a remote connection with the Raspberry Pi to manage and collect data efficiently.
\\
\\
\noindent{\bf Dataset:} 
We collected data from a classroom and a research lab. Classroom represents \textit{dense settings}, where students sit close to one another, while lab represents \textit{sparse settings}, where occupants sit or walk at a distance. In both settings, the setup was mounted on a wall, and the occupants maintained a distance from the setup. The camera's FOV captured 12 to 15 students in a classroom setup. The classroom settings included occlusion cases, varying densities, different head orientations, and students with/without glasses. While in the lab, occupants did different activities like standing, sitting, and walking with varying head poses and with/without glasses. The camera captured a maximum of 6 occupants in its FOV, showcasing different head orientations and varying distances from the camera. 

We collected 5,579 thermal images, including 2,735 from a classroom and 2,844 from a lab. The raw dataset consists of BGR thermal images with a resolution of $160 \times 120$ pixels, as captured by the FLIR thermal camera. To evaluate the generalizability and performance of our models under varying environmental conditions, we created eight different dataset combinations using the collected images from both classroom and lab environments. These combinations are detailed in Table~\ref{tab:dataset table}.

Thermal images acquired from the FLIR Lepton 3.5 camera have a native resolution of $160\times120$ pixels. While this resolution is sufficient for temperature acquisition, it poses challenges for deep learning tasks such as face detection and feature extraction. To address these issues and ensure compatibility with advanced object detection architectures, the images were resized to $640\times640$ pixels using bilinear interpolation. This resizing method balances computational efficiency with preserving visual details, minimizing aliasing artifacts and blurriness. By calculating each pixel's value as a weighted average of the four nearest neighbors, bilinear interpolation smoothens transitions and retains essential thermal features for model training.

The decision to resize to $640\times640$ was driven by the requirements of YOLO models, including YOLOv11, which are pre-trained on datasets with this resolution. Aligning the dataset with these models' input dimensions and pre-trained weights enhances their ability to accurately interpret spatial patterns and thermal features, leading to improved feature extraction and detection performance. Experimental results confirm that resizing significantly boosts model accuracy and generalizability.

In addition to resizing, data augmentation techniques such as auto-orientation and horizontal flipping were applied to expand the dataset and simulate real-world variations. These augmentations enhance the model's robustness and precision by exposing it to diverse scenarios. By providing a more comprehensive region of interest (ROI), the resized and augmented datasets improve the model's ability to generalize to thermal face detection in diverse settings.
Finally, to promote further research, we publicly release the dataset created as part of this work~\cite{indian-dataset}.
 \begin{table}[!t]
\centering
\small
%\scriptsize
\caption{Different combinations of datasets obtained from classroom and lab data.}
\begin{tabular}{lcc}
\toprule
\textbf{Dataset}                     & \textbf{Alias} & \textbf{\#} \\ \midrule
Raw classroom data                   & D1             & 2735                          \\ 
Resized classroom data               & D2             & 2735                          \\  
Resized and augmented classroom data & D3             & 4799                          \\  
Raw lab data                         & D4             & 2844                          \\ 
Resized lab data                     & D5             & 2844                          \\ 
Resized and augmented lab data       & D6             & 4962                          \\  
Resized combined data                & D7             & 5579                          \\  
Resized and augmented combined data  & D8             & 9761                          \\ \bottomrule
\end{tabular}%
\label{tab:dataset table}
%\end{wraptable}
\end{table}
     \begin{figure*}[]
            \centering
            \includegraphics[scale=0.37]{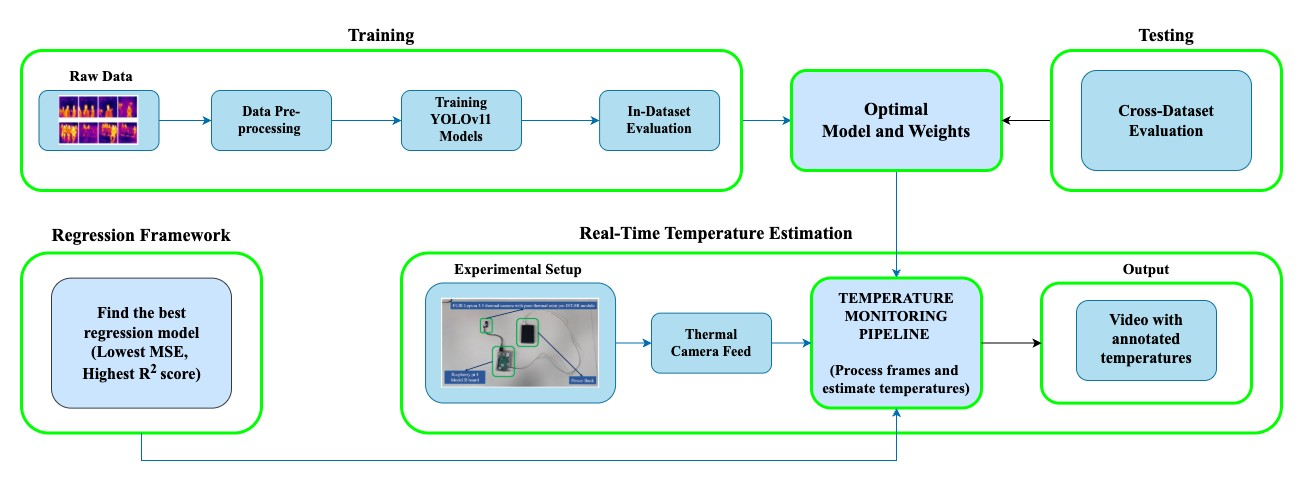}  
            \caption{Flowchart of the non-invasive temperature monitoring system.}
            \label{fig:temperature-pipeline}
    \end{figure*}
\\
\\
\noindent\textbf{Ground truth data:} We collected a separate dataset from two participants required for mapping pixel values of thermal images to body temperature. During the data collection, one participant used a 1000W hair dryer to raise his forehead temperature, while the other participant maintained a normal body temperature by remaining idle. The participants sat three meters away from the fixed setup and maintained a gap of one meter between themselves. We collected 100 images in batches of five. These batches alternated between five images with normal temperature readings, where both participants had normal body temperatures, and five images with abnormal high-temperature readings, where one participant had a normal body temperature and the other subject increased his forehead temperature. Additionally, we captured thermal images of a  water bottle to simulate low-temperature scenarios. For these images, the maximum pixel value represents the surrounding temperature, and the minimum pixel value represents the water bottle's temperature. The collected dataset has minimum, maximum, standard deviation, and mean temperatures as 25.8$^{\circ}$C, 38.8$^{\circ}$C, 2.26$^{\circ}$C, and  36.6$^{\circ}$C, respectively.

Additionally, we measured the actual (ground truth) temperatures of two participants and a water bottle using a contactless infrared thermometer (Walnut Medical Thermosure). During data analysis, we calibrated the FLIR thermal camera images by mapping pixel values to their corresponding temperature values based on the recorded ground truth measurements. This calibration was performed using the proposed regression framework, which will be detailed in the next section.
   
\section{Methodology}

Figure~\ref{fig:temperature-pipeline} shows a flowchart of our methodology for estimating face temperature from thermal feeds. Following is a description of the different modules present within the system. 

\subsection{Training \& Testing}

Initially, we pre-processed the collected thermal images by constructing rectangular bounding boxes (bbox) around the faces of people using the LabelImg tool\footnote{\url{https://github.com/heartexlabs/labelImg}}. These bounding boxes became our ROI. Moreover, each image's bounding box labels (i.e., bbox coordinates) were converted into the YOLO format required for training the YOLO models. Next, we followed two different evaluation strategies: (i) In-dataset evaluation and (ii) Cross-dataset evaluation. In In-dataset, the same dataset was used for training and testing, while in cross-dataset evaluation, different datasets were used for training and testing purposes. Following are the combinations of the cross-dataset evaluations: (i) \textbf{C1:} classroom data  X lab data, (ii) \textbf{C2:} classroom data X  combined data, (iii) \textbf{C3:} lab data X  classroom data
(iv) \textbf{C4:} lab data X combined data,  (v) \textbf{C5:} combined data X classroom data, and (vi) \textbf{C6:} combined data  X  lab data.  In the mentioned combinations, the notation $Di$ X $Dj$ indicates that the best model trained on the $Di$ data is later used for conducting inference on the validation set of the $Dj$ data. The cross-dataset evaluation is important to check the model's robustness and generalizability. 

Finally, the best and last weights of all the trained models were stored for inferencing faces from thermal images, videos, and live thermal camera feeds.

\begin{table*}[!t]
\scriptsize
\centering
\caption{Training results on the classroom data for various parameter configurations. Common parameters include: Epochs- 100, Batch size - 16, Image size - 640 x 640, and Optimizer - SGD. Bold values represent the best results.}
\label{tab:train_tab1}
\begin{tabular}{ccccccc}
\toprule
\textbf{Model} & \textbf{Parameters} & \textbf{Dataset} & \textbf{Precision} & \textbf{Recall} & \textbf{mAP0.5} & \textbf{mAP0.5:0.95} \\ \midrule
\textbf{YOLOv11s}                     & Image size- 120x120 & D1               & 84.8               & 83.3            & 84.1            & 32.7                 \\
\textbf{YOLOv11n}                     & Image size- 120x120 & D1               & 84.3               & 83.0            & 83.9            & 31.7                 \\
\textbf{YOLOv11s}                     & -                   & D2               & 87.7               & 88.2            & 88.5            & 35.6                 \\
\textbf{YOLOv11n}                     & -                   & D2               & 87.3               & 88.1            & 88.8            & 36.3                 \\
\textbf{YOLOv11s}                     & -                   & D3               & 89.9               & 91.1            & 91.2            & 39.9                 \\
\textbf{YOLOv11n}                     & \textbf{-}          & \textbf{D3}      & \textbf{89.7}      & \textbf{91.2}   & \textbf{91.5}   & \textbf{40.3}        \\
\textbf{YOLOv11s}                     & Epochs- 200         & D3               & 90.7               & 91.7            & 90.7            & 43.3                 \\
\textbf{YOLOv11n}                     & Epochs- 200         & D3               & 90.5               & 91.7            & 90.5            & 45.3                 \\
\textbf{YOLOv11s}                     & Optimizer- Adam     & D3               & 88.3               & 90.1            & 89.1            & 35.5                 \\
\textbf{YOLOv11n}                     & -                   & D3               & 90.3               & 91.4            & 91.1            & 39.3                 \\
\textbf{YOLOv11s (15 layers frozen)}  & -                   & D3               & 88.7               & 88.9            & 88.5            & 35.8                 \\
\textbf{YOLOv11s (10 layers frozen )} & -                   & D3               & 88.9               & 90.6            & 89.8            & 36.8                 \\
\textbf{YOLOv11s (5 layers frozen)}   & -                   & D3               & 90.2               & 91.5            & 91.0            & 39.3            \\ \bottomrule    
\end{tabular}
\end{table*}

\subsection{Regression Framework}

The following equation represents a simple linear regression model for mapping pixel to temperature values. 
\begin{equation}
Temperature = \beta_0 + \beta_1 * Pixel\_Value + \epsilon
\label{equ: eq1}
\end{equation}

Where $Temperature$ represents ground truth temperature and $Pixel\_Value$ is the maximum pixel value within the ROI. The equation might change depending on the regression model used to predict the temperature. The maximum pixel value was used to estimate the highest temperature within the ROI, thus enabling the identification of individuals exhibiting unusual or elevated body temperatures. 

We evaluated 11 different regression models for estimating temperature: Linear, Ridge, Lasso, Elastic net, Support vector, K Neighbors, Gaussian process, Decision tree, Gradient Boosting, Random Forest, and XGBoost. The models were evaluated using Mean Squared Error (MSE) and $R^2$ score defined as follows
 \begin{equation}
MSE = \frac{1}{n}\sum_{i=1}^{n}(T_i - \hat{T_i})^2
 \end{equation}

  \begin{equation}
R^2 = 1 - \frac{\sum_{i=1}^{n}(T_i - \hat{T_i})^2}{\sum_{i=1}^{n}(T_i - \bar{T})^2}
 \end{equation}

 Where $T$, $\hat{T}$  represents ground truth and the regressor estimated temperature, respectively. $\bar{T}$ is the mean temperature of all the  $n$ readings used for temperature estimation. An ideal regression model should result in the highest $R^2$ score and the minimum MSE. Optimal parameters of regression models were obtained with cross-validation and hyperparameter tuning. 
The framework resulted in the best regression model with estimated coefficients ($\beta_0$, $\beta_1$) and loss ($\epsilon$).

\subsection{Real-time Temperature Estimation}

We transferred the best thermal face detection and regression models to a Raspberry Pi. We also transferred the best weights obtained from the most optimal model for thermal face detection to the Raspberry Pi, enabling real-time temperature monitoring using live feeds captured by a thermal camera. 

\noindent\textbf{Temperature Monitoring Pipeline:}
The temperature monitoring pipeline was meticulously designed to streamline the process, ensuring its adaptability to any dataset and experimental setup. It employs the most optimal models from the regression framework and thermal face detection, along with their corresponding weights, and utilizes the live camera feed as input to estimate temperatures based on the maximum pixel value within the ROI.

The pipeline commences with the input thermal video undergoing essential pre-processing steps. These include: (i) loading the image, video, or live feed, (ii) filling empty frames (without any person) with null labels, and (iii) pairing the corresponding thermal image frames with their respective bounding box (bbox) labels. Additionally, BGR frames are converted to grayscale for consistent processing.

The YOLO model's detect script is employed for facial region detection, leveraging the optimal model's best weights to predict bbox around facial regions. These predicted YOLO bbox labels are then denormalized to their original bbox labels. The next step involves cropping the ROI from each frame to extract the maximum pixel value. A minimum thresholding area is implemented for each bbox to address edge cases involving excessively small bounding boxes. This ensures more robust temperature estimation in dense scenarios. The temperature labels of the ROI are estimated using the best regression model obtained from the regression framework. 

These estimated temperature labels are then added to each frame, enabling the generation of a new output video. The new output video showcases both bounding boxes and corresponding temperature labels for each individual in the frame. 

Finally, we get a live video feed with bbox around the facial regions and their estimated temperatures overlaid on each bbox.

\begin{table*}[!t]
\centering
\scriptsize
\caption{Training results on the lab data for various parameter configurations. Common parameters include: Epochs - 100, Batch size - 16, Image size - 640 x 640, optimizer - SGD. Bold values represent best results.}
\label{tab:train_tab2}
\begin{tabular}{ccccccc}
\toprule
\textbf{Model} & \textbf{Parameters} & \textbf{Dataset} & \textbf{Precision} & \textbf{Recall} & \textbf{mAP0.5} & \textbf{mAP0.5:0.95} \\ \midrule
\textbf{YOLOv11n}                     & Image size- 120x120                     & D4          & 84.8          & 84.5          & 88.3          & 36.5          \\
\multicolumn{1}{l}{\textbf{YOLOv11s}} & \multicolumn{1}{l}{Image size- 120x120} & D4          & 86.5          & 85.0          & 88.8          & 37.1          \\
\textbf{YOLOv11n}                     & -                                       & D5          & 88.3          & 88.0          & 91.3          & 38.5          \\
\multicolumn{1}{l}{\textbf{YOLOv11s}} & -                                       & D5          & 89.4          & 87.8          & 91.5          & 38.7          \\
\textbf{YOLOv11n}                     & \textbf{-}                              & \textbf{D6} & \textbf{91.5} & \textbf{92.3} & \textbf{94.5} & \textbf{45.6} \\
\textbf{YOLOv11s}                     & -                                       & D6          & 92.3          & 93.0          & 94.2          & 45.4          \\ \bottomrule
\end{tabular}
\end{table*}

\begin{table*}[!t]
\centering
\scriptsize
\caption{Training results on combined classroom and lab data for various parameter configurations. Common parameters include: Batch size - 8, Image size - 640 x 640, optimizer - SGD. Bold values represent best results.}
\label{tab:train_tab3}
\begin{tabular}{ccccccc}
\toprule
\textbf{Model} & \textbf{Parameters} & \textbf{Dataset} & \textbf{Precision} & \textbf{Recall} & \textbf{mAP0.5} & \textbf{mAP0.5:0.95} \\ \midrule
\textbf{YOLOv11n} & Epochs- 20  & D7 & 88.2 & 90.3 & 89.2 & 37.6 \\
\textbf{YOLOv11n} & Epochs- 100 & D7 & 91.2 & 91.5 & 92.5 & 48.6 \\
\textbf{YOLOv11s} & Epochs- 100          & D7               & 92.2      & 92.2   & 93.2   & 49.6        \\
\textbf{YOLOv11n} & \textbf{Epochs- 100} & \textbf{D8}      & \textbf{94.2}      & \textbf{95.5}   & \textbf{94.3}   & \textbf{50.2}        \\
\textbf{YOLOv11s} & Epochs- 100 & D8 & 95.2 & 96.5 & 94.2 & 50.1
\\ \bottomrule
\end{tabular}
\end{table*}

\begin{table*}[]
\centering
%\scriptsize
\caption{Cross-dataset evaluation results obtained using best models (YOLOv11s (classroom data), YOLOv11s (lab data), YOLOv11n (combined data)) trained on different datasets. Bold values represent best results.}
\label{tab:cross-dataset-results}
\resizebox{\textwidth}{!}{%
\begin{tabular}{cccccccccccccc}
\toprule
\textbf{Validation} &
  \textbf{Type} &
  \multicolumn{4}{c}{\textbf{Train on classroom data}} &
  \multicolumn{4}{c}{\textbf{Train on lab data}} &
  \multicolumn{4}{c}{\textbf{Train on combined data}} \\ \cmidrule(lr){3-6} \cmidrule(lr){7-10} \cmidrule(lr){11-14} 
\textbf{Dataset} &
   &
  \textbf{Precision} &
  \textbf{Recall} &
  \textbf{mAP0.5} &
  \textbf{mAP0.5:0.95} &
  \textbf{Precision} &
  \textbf{Recall} &
  \textbf{mAP0.5} &
  \textbf{mAP0.5:0.95} &
  \textbf{Precision} &
  \textbf{Recall} &
  \textbf{mAP0.5} &
  \textbf{mAP0.5:0.95} \\ \hline
\textbf{Lab Data}       & \textbf{D4} & 79.1 & 68.1 & 60.1 & 28.9 & -    & -    & -    & -    & 89.8 & 81.5 & \textbf{88.2} & 36.8 \\
                        & \textbf{D5} & 79.4 & 62.2 & 69.5 & 38.9 & -    & -    & -    & -    & 91.7 & 91.4 & \textbf{94.1} & 40.6 \\
                        & \textbf{D6} & 80.2 & 68,3 & 67.8 & 30.4 & -    & -    & -    & -    & 91.3 & 92.4 & \textbf{94.3} & 42.3 \\ \hline
\textbf{Classroom Data} & \textbf{D1} & -    & -    & -    & -    & 68.4 & 47.5 & 45.4 & 20.4 & 82.3 & 84.8 & \textbf{87.8} & 35.6 \\
                        & \textbf{D2} & -    & -    & -    & -    & 72.9 & 68.8 & 65.5 & 27.6 & 92.3 & 92.1 & \textbf{95.2} & 42.5 \\
                        & \textbf{D3} & -    & -    & -    & -    & 72.7 & 69.0 & 66.7 & 27.3 & 90.3 & 91.5 & \textbf{94.5} & 32.2 \\ \hline
\textbf{Combined data}  & \textbf{D7} & 85.8 & 84.5 & 88.2 & 40.2 & 83.6 & 75.1 & 75.7 & 29.8 & -    & -    & -             & -    \\
                        & \textbf{D8} & 88.2 & 88.7 & 89.6 & 41.2 & 85.8 & 77.6 & 78.3 & 30.9 & -    & -    & -             & -    \\ \bottomrule
\end{tabular}%
}
\end{table*}

\subsection{Experimental settings \& Evaluation metrics}

We trained our models on YOLOv11\cite{khanam2024yolov11}, its lightweight variants, YOLOv11n (nano). The lightweight variant is suitable for deployment on edge devices like Raspberry Pi. We also used the YOLOv11s for training to assess performance discrepancies between the lightweight and the larger model. YOLOv11s's frozen layers effectively reduced the number of parameters and size of the models, facilitating comparison with the lightweight models. We gradually unfroze the layers until the entire model is unfrozen. The unfreezing operation allowed us to observe the impact of increasing parameters and the model's overall size, as shown in Table \ref{tab:train_tab1}. 

The models' performance was assessed based on the mAP score, where a higher score indicates better performance. The metrics include precision (TP/TP+FP), recall (TP/TP+FN), mAP@0.5, and mAP@0.5:0.95 scores. The mAP@0.5 score and the mAP@0.5:0.95 scores indicate the mean average precision at an intersection over union (IOU) threshold (IOU = area of overlap / total area of union) of 0.5 and a range of variable IOU thresholds from 0.5 to 0.95, respectively. The mAP score represents the area under the precision-recall curve, averaged over all the classes in the dataset.

\textit{Note:} We could not evaluate our methods on existing datasets as there are no publicly available datasets with dense settings.

\section{Results}
 In this section, we will present the face detection results (In and Cross-dataset) and the temperature monitoring results.

   \begin{figure*}[]
    \centering
    % \resizebox{0.90\columnwidth}{!}{
    \begin{subfigure}[b]{0.49\textwidth}
        \centering
        \includegraphics[width=\textwidth]{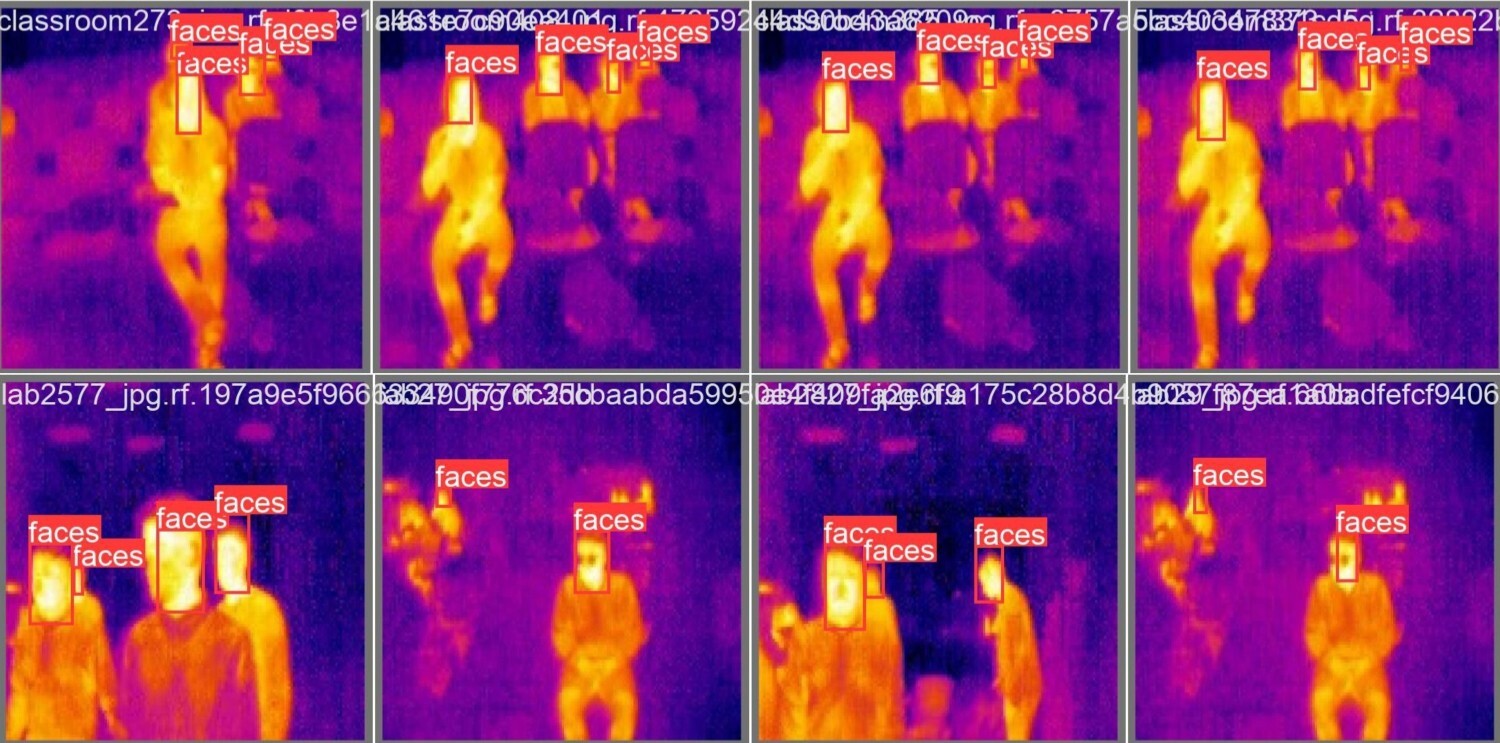}
        \caption{Actual bounding box labels}
        \label{fig:face_real}
    \end{subfigure}
    \hfill
    \begin{subfigure}[b]{0.49\textwidth}
        \centering
        \includegraphics[width=\textwidth]{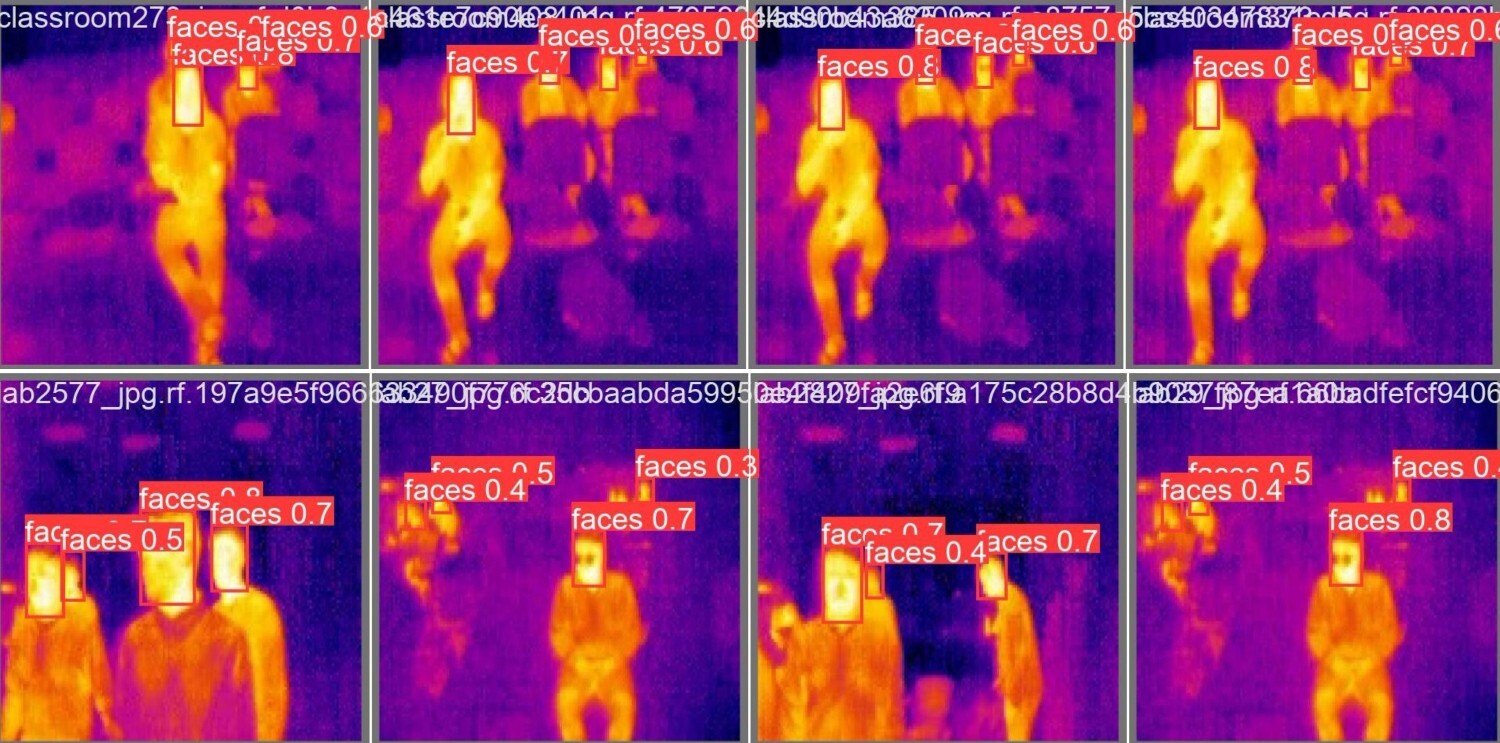}
        \caption{Predicted bounding box labels}
        \label{fig:face_pred}
    \end{subfigure}%
    % }
    \caption{Actual and predicted bounding boxes for classroom and lab datasets. Numbers associated with the ``faces" label in the predicted bounding boxes show the probability of a face captured within the bounding box.}
    \label{fig:face detection}
    \end{figure*}

\subsection{Thermal face detection results}
\noindent\textbf{In-dataset evaluation:}
Tables~\ref{tab:train_tab1}, \ref{tab:train_tab2}, and \ref{tab:train_tab3} show the results obtained during training on the validation data of classroom, lab, and combined (both classroom and lab) datasets, respectively, using various parameter configurations and models. Table~\ref{tab:train_tab1} shows that the YOLOv11n model, trained on the D3 dataset (i.e., resized and augmented classroom dataset), emerges as the best model for the classroom data. This model was trained for 100 epochs, utilizing a batch size of 16 and the SGD optimizer. 
Table~\ref{tab:train_tab2} shows that the best model for the lab data is the YOLOv11n model. This model was trained for 100 epochs, employing a batch size of 16 and the SGD optimizer. It was trained on the D5 dataset (i.e., resized and augmented lab dataset). Table \ref{tab:train_tab3} shows the best model for the combined data is the YOLOv11n model. This model was trained for 100 epochs, utilizing a batch size of 8 and the SGD optimizer. It was trained on the  D7 dataset (i.e., resized and augmented combined dataset).\\

\noindent\textbf{Cross-dataset evaluation:}
Table~\ref{tab:cross-dataset-results} shows the cross-dataset evaluation results for lab, classroom, and combined data. In this evaluation, the bounding boxes around the faces in the validation sets were predicted using the best weights obtained from the training models on lab, classroom, and combined data. Results show that the model trained on the combined dataset outperformed the models trained on the individual classroom and lab datasets. This signifies the model's capacity to generalize well to unseen data.

    \begin{figure} 
        \centering
             \includegraphics[scale=0.25]{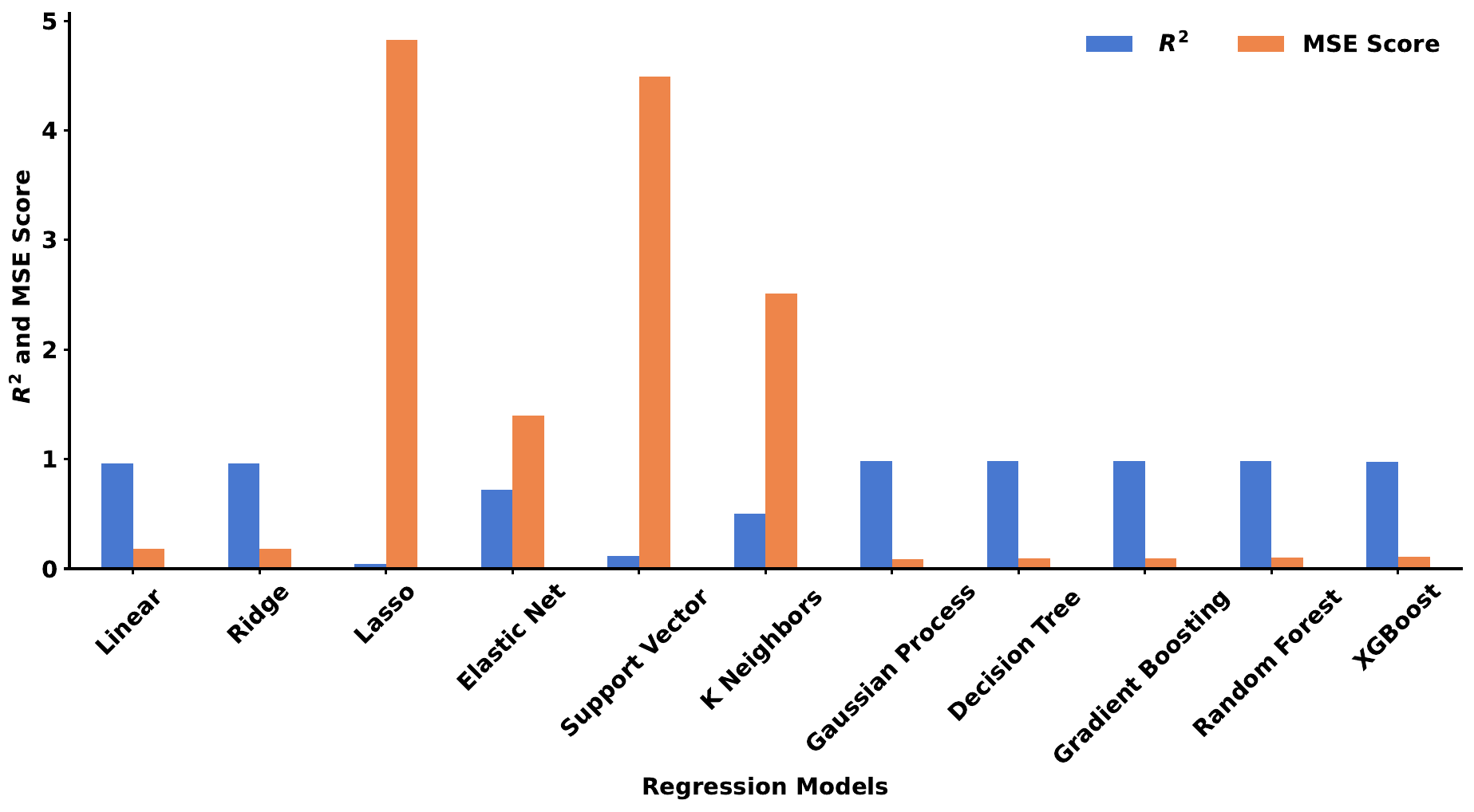}  
        \caption{Performance of different regression models on ground truth lab data with abnormal temperature changes.}
        \label{fig:reg}
    \end{figure}

In and cross-dataset evaluation results revealed that the best-performing model was trained on the resized and augmented combined data, employing 100 epochs, a batch size of 8, and the SGD optimizer. It exhibited a remarkable mAP@0.5 score of 94.3, shown in Table \ref{tab:train_tab3} while keeping a mAP@0.5 score of above 87.8 for all the cross-dataset combinations, while achieving a maximum map@0.5 score of 95.2 as shown in Table \ref{tab:cross-dataset-results}. This indicates the robust ability of the model to detect thermal faces in varying environments and conditions.

Figure~\ref{fig:face detection} shows actual and predicted bounding boxes obtained with the best model (YOLOv11n) on test data. Predicted bounding boxes with probability scores show the probability of capturing a face correctly within the bounding box. 

    \begin{figure*}[!t]
    \centering
    \begin{subfigure}[b]{0.49\textwidth}
        \centering
        \includegraphics[width=\textwidth]{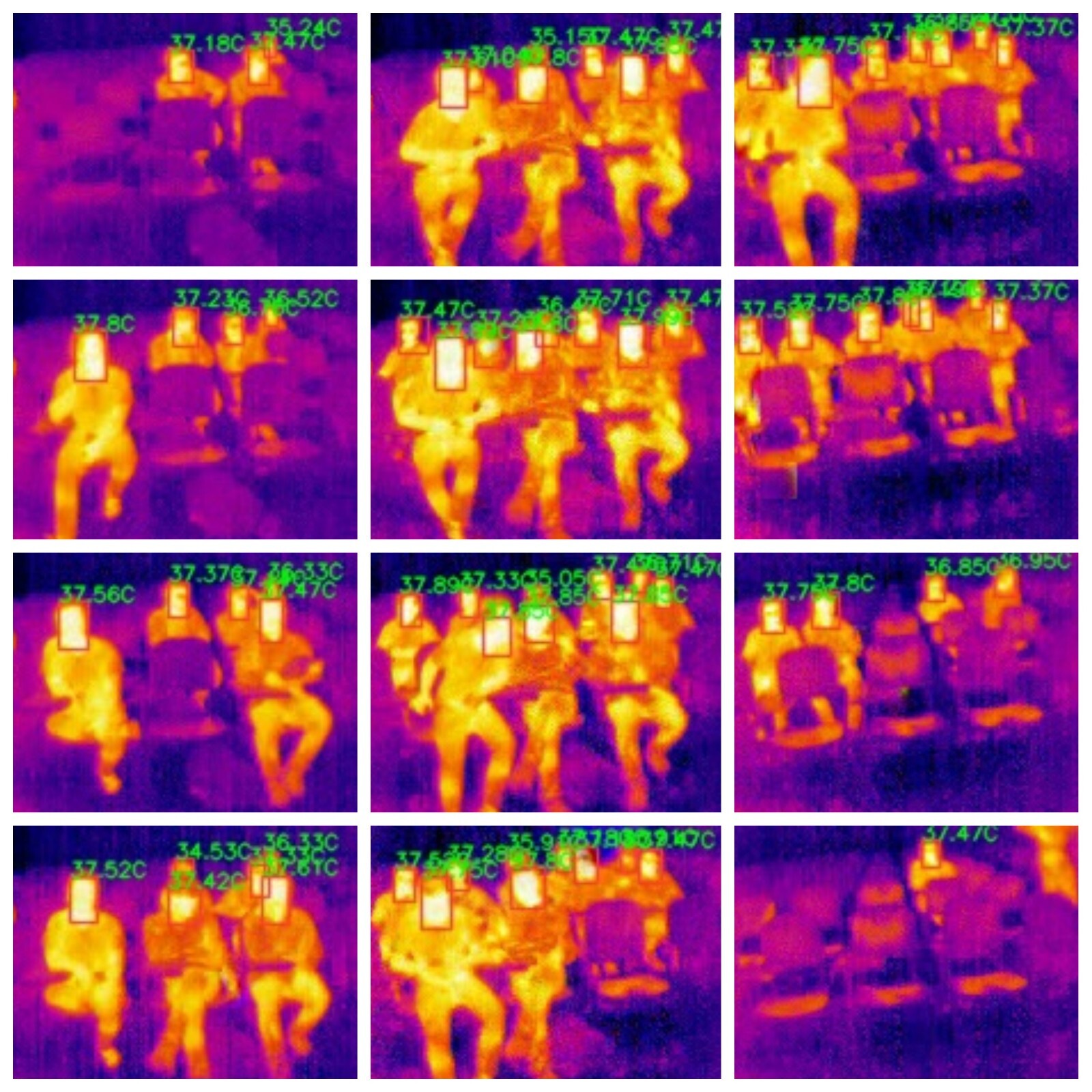}
        \caption{Classroom settings}
        %\label{fig:sub1}
    \end{subfigure}
    \hfill
    \begin{subfigure}[b]{0.49\textwidth}
        \centering
        \includegraphics[width=\textwidth]{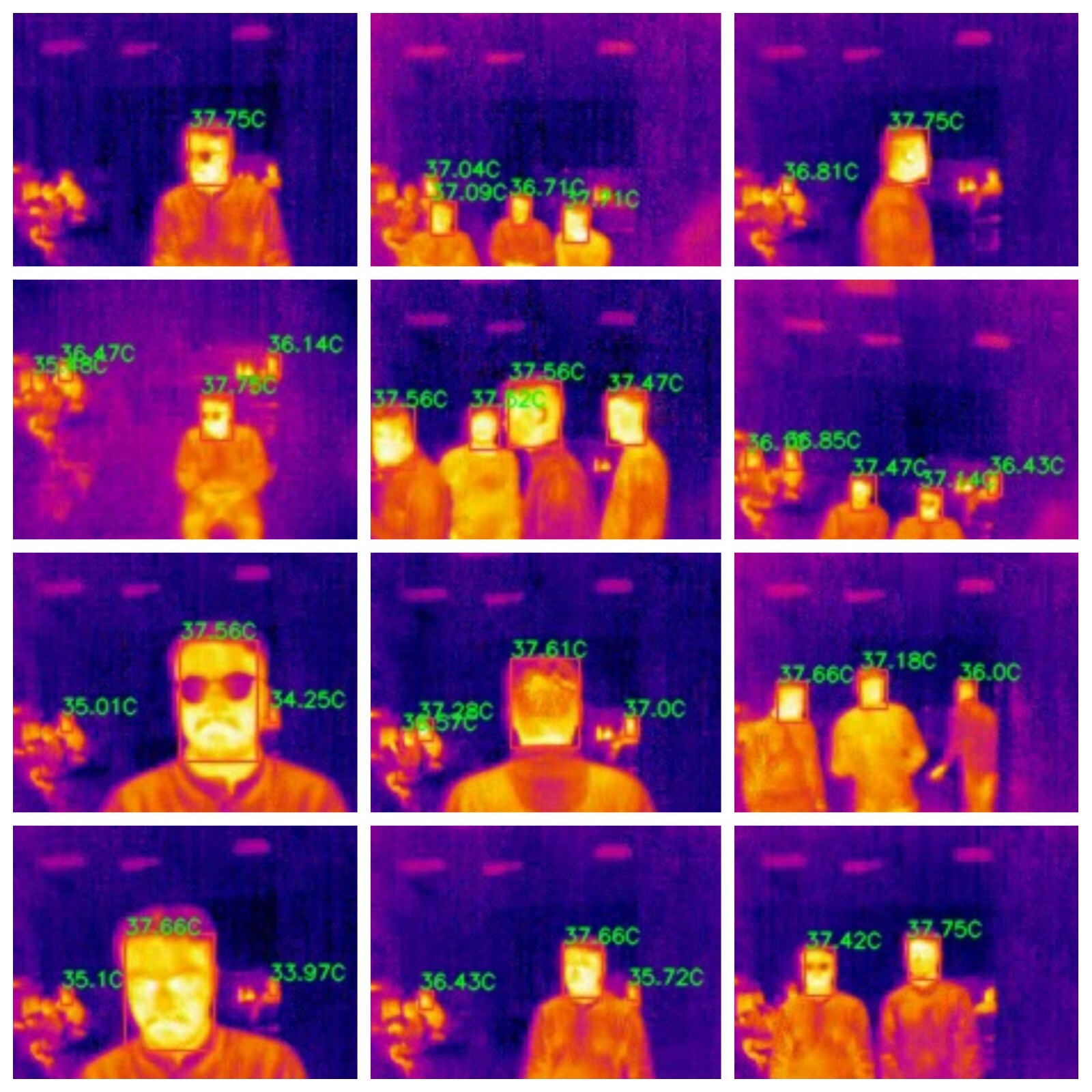}
        \caption{Lab settings}
        %\label{fig:sub2}
    \end{subfigure}
    \caption{Temperature monitoring system in action: Predicted bounding boxes and estimated temperatures in different settings on test datasets.}
    \label{fig:bbox-temperature-predicions}
    \end{figure*}
\subsection{Temperature monitoring results}
 
Figure \ref{fig:reg} shows the $R^2$ score and MSE of different regression models. Gaussian process regressor emerged as the most proficient regression model for the captured data, with the highest $R^2$ score and the lowest MSE. However, when we tested it on the classroom and lab data, it predicted temperatures higher than 38$^{\circ}$C, which does not align with the actual temperatures as nobody in the classroom or lab had a fever or any medical condition having elevated body temperature. As a result, we opted for ridge regression. This model exhibited a reasonable $R^2$ score and MSE. We applied Ridge regression to the captured images, videos, and live thermal camera feeds to predict the temperature of each individual within the frame. Figure \ref{fig:bbox-temperature-predicions} shows some classroom and lab images with bbox labels and temperatures predicted in real-time. This model enabled real-time temperature estimation without introducing any lag in the original live video feed.

\section{Discussion}

\subsection{Thermal face detection}

A thorough analysis of both In-dataset and Cross-dataset evaluation showed that the YOLOv11n model trained on the combined dataset outperformed other models. This is because it was trained on the combined dataset with both the lab and classroom data. It also captured the effects of varying distances from the experimental setup, densities, people postures, head orientations, and conditions. This lightweight model boasts fewer parameters, rendering it highly suitable for deployment in low-hardware power setups, such as our experimental setup. Through extensive training, we ensured that the model could handle challenging scenarios, including occlusion, individuals wearing glasses, and completely dark environments. As a result, our trained models achieved precise ROI predictions on unseen real-world data.

\subsection{Real-world challenges}

Our study addresses key real-world challenges to ensure robust thermal imaging and analysis. \textbf{Camera calibration} was meticulously conducted using FLIR's proprietary software, adjusting parameters like emissivity, reflective temperature, and target distance to ensure accurate and reliable temperature measurements across diverse environments. \textbf{Lighting conditions}, often a limitation in standard imaging, were inherently mitigated as the FLIR thermal camera operates on infrared radiation, maintaining consistent performance in complete darkness or overexposed settings. Additionally, we prioritized \textbf{ethical considerations} by leveraging thermal imaging's non-invasive nature, which captures only heat signatures, preserving privacy and avoiding collecting personally identifiable information. All data collection adhered to strict ethical protocols, with informed written consent obtained from participants, ensuring transparency and compliance with professional standards.

   \begin{figure*}[!t]
    \centering
    
    \begin{subfigure}[b]{0.60\textwidth}
        \centering
        \includegraphics[scale=0.35]{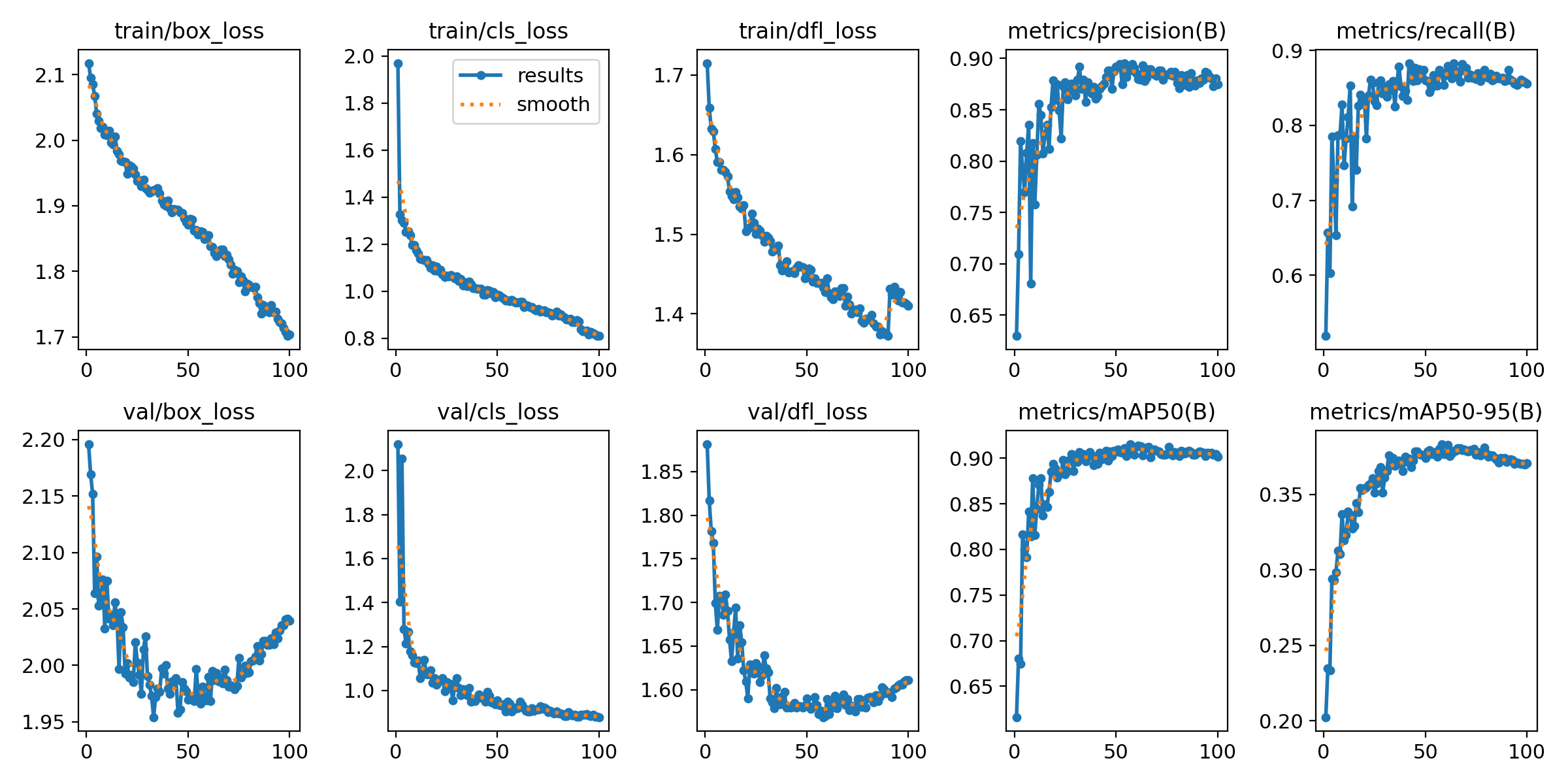}
        %\caption{loss, precision and recall metrics vs number of epochs for the best model}
        \caption{Loss (training \& validation), Precision, Recall, mAP\_0.5, and mAP\_0.5:0.95 of the best model. X-axis in each plot represents epochs (\#) and Y-axis represents metric value.}
        \label{fig:metrics}
    \end{subfigure}
    \hfill
    \begin{subfigure}[b]{0.35\textwidth}
        \centering
        \includegraphics[scale=0.23]{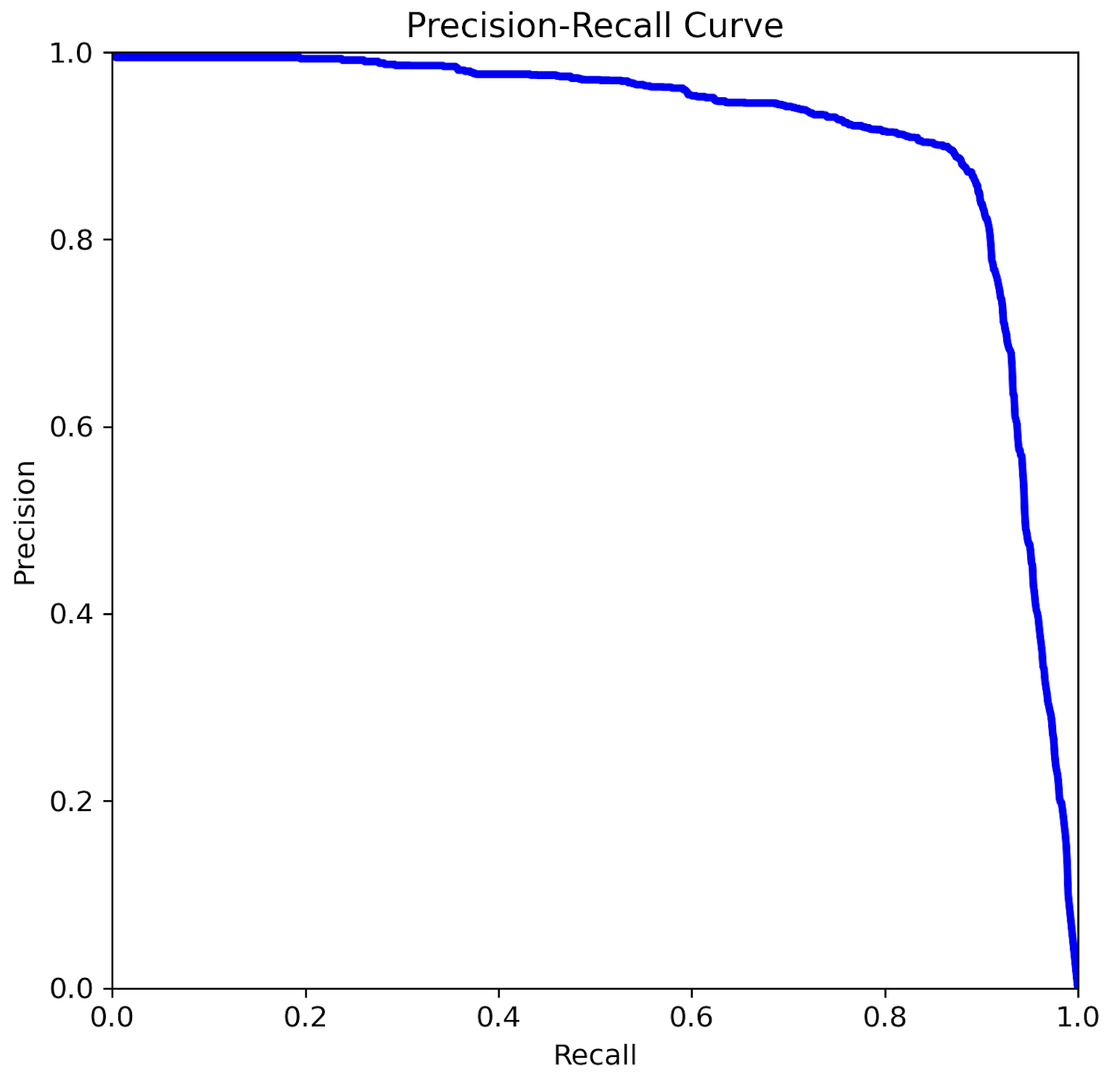}
        \caption{Precision Recall curve of the best model}
        \label{fig:precision-recall-curve}
    \end{subfigure}
    \label{fig:precision-recall-all-results}
    \caption{Different metrics of the best model (YOLOv11n).}
    \end{figure*}

\subsection{Temperature monitoring pipeline \& Regression framework}

The developed temperature monitoring pipeline demonstrates robustness in handling diverse datasets and experimental setups, making it adaptable to variable scenarios. To build the regression framework, we collected a specific dataset from a lab setting involving two subjects: one with a normal and another with an abnormally high temperature. By capturing an image that included both individuals in the same frame, we could observe the relative change in pixel values between the high-temperature and normal-temperature subjects. This approach was necessary due to the characteristics of thermal images represented in the BGR format, which maps temperature values within the range of 0-255, regardless of the actual temperature magnitude. By analyzing the pixel value variations in the presence of both subjects, we gained insights into accurately estimating temperatures in such scenarios.

Among the regression models evaluated, the Gaussian Process Regressor achieved the best performance in terms of Mean Squared Error (MSE) and $R^2$ scores. However, it was not used for temperature estimation due to its tendency to predict abnormally high temperatures for certain subjects. This behavior was likely caused by overfitting on the training dataset and the model's limited generalization to unseen scenarios.

Other models, such as Decision Tree, Gradient Boosting, Random Forest, and XGBoost regressors, demonstrated performance comparable to the Gaussian Process Regressor in terms of MSE and $R^2$. However, these models were computationally intensive and unsuitable for real-time temperature estimation in live video feeds. Their predictions introduced delays in processing successive video frames, resulting in time lags and reduced practicality in the experimental setup.

To overcome these limitations, we opted for Ridge Regression, a linear regression model with an L2 regularization term in its loss function. Ridge Regression effectively mitigated overfitting while achieving reasonable MSE and $R^2$ scores compared to the Gaussian Process Regressor. Additionally, it consistently predicted temperature values that closely matched the actual human body temperatures recorded in the ground truth dataset. The choice of Ridge Regression was further motivated by its computational simplicity and ability to support real-time temperature estimation, addressing the challenges posed by more complex models like Random Forest, Decision Tree, Gradient Boosting, and XGBoost regressors.

\subsection{Robust face detection}
Robust face detection was a central focus of our implementation, achieved through advanced algorithms and rigorous manual data annotation. The annotation process emphasized cases with severe occlusions to ensure the dataset captured diverse real-world scenarios, enhancing the model’s robustness. We utilized a fine-tuned version of the YOLOv11 model, which demonstrated effective face detection even under challenging conditions, such as significant occlusions and varying thermal signatures. Data augmentation techniques were employed during training to bolster the model’s resilience to these challenges. Furthermore, lighting variations were inherently addressed by the FLIR thermal infrared camera, which provides consistent imaging regardless of ambient light.

 \subsection{Sensitivity analysis}
We analyzed the optimal model by examining the effects of varying epochs, optimizers, and batch sizes. Our primary evaluation metric was mAP@0.5 scores. Increasing the number of epochs positively impacted both the mAP@0.5 score and mAP@0.5:0.95, as depicted in Figure \ref{fig:metrics}. However, once we surpassed 100 epochs, there was no significant improvement in the mAP@0.5 score, as indicated in Table \ref{tab:train_tab1}. This suggests that the results reach a saturation point after 100 epochs.

 Regarding optimizers, models trained with SGD outperformed those trained with Adam, as shown in Table \ref{tab:train_tab1}. Interestingly, the small-sized models did not exhibit a noteworthy enhancement in mAP@0.5 scores. This implies that the nano model can be used for thermal face detection with comparable efficiency to the small model, with fewer parameters, lower computational costs, and shorter evaluation time. Precision-recall analysis (Figure \ref{fig:metrics}) revealed that the area under the curve increased with more epochs, positively impacting the mAP scores. Additionally, the object loss and box loss decreased with each epoch.

Furthermore, we comprehensively compared regression models, incorporating hyperparameter tuning (such as K value, C, epsilon, max\_depth, n\_estimators, criterion, etc.). This allowed us to derive the best parameter configurations for the regression models, resulting in superior temperature estimation from pixel values in thermal images, videos, and live thermal camera feeds.

 \subsection{Future Work}

Future research will focus on expanding the dataset to include diverse environmental conditions, demographics, and thermal scenarios. This expansion will evaluate the generalizability of the models and ensure their applicability across various real-world situations. Incorporating a comprehensive dataset will also enable the models to capture a broader range of thermal patterns and improve detection accuracy under varying conditions.

To further enhance detection capabilities, we plan to explore advanced object detection architectures, including state-of-the-art (SOTA) models, specifically tailored for thermal imaging. These architectures will be designed to address challenges such as severe occlusions and varying thermal signatures. Additionally, domain adaptation, few-shot learning, and self-supervised learning techniques will be investigated to improve model robustness in diverse and unforeseen scenarios.

Another area of focus will be addressing the impact of occlusions through multi-camera setups. By capturing subjects from multiple angles, these setups will minimize the effects of occlusions on temperature estimation and detection accuracy. Additionally, hardware optimization will be explored to make the system more energy-efficient and scalable. Investigating simple thermography models for real-time, computationally inexpensive temperature estimation will further extend the system's utility and accessibility.

In addition to the avenues discussed earlier, the evolution of robotic-assisted surgical systems, as detailed by \cite{alsajri2024future} underscores the transformative potential of precision, real-time monitoring in clinical applications. In a similar vein, future iterations of our temperature monitoring system could leverage advanced sensor fusion techniques and adaptive control mechanisms—approaches that have already demonstrated success in robotic surgery—to further enhance accuracy and responsiveness. By drawing on these innovations, our system can be optimized to better serve dynamic and congested environments, ensuring more reliable and actionable health monitoring in real-world settings.

\section{Conclusion}

This work introduced a novel non-invasive temperature monitoring system with several distinctive aspects, including
(i) The system operates on an edge device, eliminating the need for computational power.
It uses a lightweight YOLOv11n model for face detection and then combines that with the regression framework for estimating the temperature of each individual in real-time.
(ii) It is a stand-alone system; however, it offers the possibility of creating a distributed network of similar edge devices to monitor large areas or buildings through a unified dashboard. 
(iii) The system is portable, exhibiting superior performance in sparse and dense settings. Additionally, it can be deployed on a wall and powered by various sources, such as a power bank. 

The system was evaluated on a diverse thermal image dataset.
As no publicly available dataset with dense settings is present, we made the dataset publicly accessible. We aim to encourage and facilitate further research by releasing the dataset.

\balance
\bibliographystyle{IEEEtran}
\bibliography{citations.bib}

\end{document}